\documentclass{article}

\usepackage{hyperref}       
\usepackage{url}            
\usepackage{graphicx}
\usepackage{subfigure}
\usepackage{caption}
\usepackage{subfiles}
\usepackage{tabularx}
\usepackage{etoolbox}
\usepackage[T1]{fontenc}
\usepackage[dvipsnames]{xcolor}
\usepackage{bbm}
\usepackage{multirow}
\usepackage{enumitem}
\usepackage{amssymb}
\usepackage{amsmath}
\usepackage{amsthm}
\usepackage{xspace}
\usepackage{float}
\usepackage{booktabs}       
\usepackage{amsfonts}       
\usepackage{nicefrac}       
\usepackage{microtype}      
\usepackage{color, soul}
\usepackage{titling}
\usepackage{setspace}
\newtoggle{arxiv}
\toggletrue{arxiv}

\newcommand{\yell}[1]{{\color{black}#1}}

\newcommand{\name}{\textsc{Bootleg}\xspace}

\newcommand{\para}[1]{\vspace{0.08in}\noindent\textbf{#1 }}

\iftoggle{arxiv}{
    \usepackage[numbers,sort]{natbib}
    \setlength{\textwidth}{6.5in}
    \setlength{\textheight}{9in}
    \setlength{\oddsidemargin}{0in}
    \setlength{\evensidemargin}{0in}
    \setlength{\topmargin}{-0.5in}
    \newlength{\defbaselineskip}
    \setlength{\defbaselineskip}{\baselineskip}
    \setlength{\marginparwidth}{0.8in}
}

\newcolumntype{Y}{>{\hsize=.7\hsize}X}
\newcolumntype{Z}{>{\hsize=1.3\hsize}X}

\makeatletter
\def\@copyrightspace{\relax}
\makeatother

\makeatletter
\def\@myauthornotes{}
\def\myauthornote#1{%
  \if@ACM@anonymous\else
    \g@addto@macro\addresses{}%
    \g@addto@macro\@myauthornotes{%
      \stepcounter{footnote}\footnotetext{#1}}%
  \fi}
\makeatother

\iftoggle{arxiv}{
    \title{Bootleg: Chasing the Tail with Self-Supervised \\Named Entity Disambiguation}
    \usepackage{authblk}
    \author[$\dagger$]{Laurel Orr}
    \author[$\dagger$]{Megan Leszczynski}
    \author[$\dagger$]{Simran Arora}
    \author[$\dagger$]{Sen Wu}
    \author[$\dagger$]{Neel Guha}
    \author[$\ddagger$]{Xiao Ling}
    \author[$\dagger$]{Christopher R{\'e}}
    \affil[$\dagger$]{Stanford University} \affil[$\ddagger$]{Apple\vspace{4pt}}
    \affil[ ]{\texttt{\{lorr1,mleszczy,simran,senwu,nguha,chrismre\}@cs.stanford.edu, xiaoling@apple.com}}
}

\predate{}
\postdate{}
\date{}
\begin{document}

\maketitle

\begin{abstract}
\label{sec:abstract}
 A challenge for named entity disambiguation (NED), the task of mapping textual mentions to entities in a knowledge base, is how to disambiguate entities that appear rarely in the training data, termed {\em tail} entities. Humans use subtle reasoning patterns based on knowledge of entity facts, relations, and types to disambiguate unfamiliar entities. Inspired by these patterns, we introduce \name, a self-supervised NED system that is explicitly grounded in reasoning patterns for disambiguation. We define core reasoning patterns for disambiguation, create a learning procedure to encourage the self-supervised model to learn the patterns, and show how to use weak supervision to enhance the signals in the training data. Encoding the reasoning patterns in a simple Transformer architecture, \name meets or exceeds state-of-the-art on three NED benchmarks. We further show that the learned representations from \name successfully transfer to other non-disambiguation tasks that require entity-based knowledge: we set a new state-of-the-art in the popular TACRED relation extraction task by 1.0 F1 points and demonstrate up to 8\% performance lift in highly optimized production search and assistant tasks at a major technology company.
\end{abstract}

\section{Introduction}
\label{sec:introduction}

Knowledge-aware deep learning models have recently led to significant progress in fields ranging from natural language understanding \cite{knowbert, poerner1911bert} to computer vision \cite{zhu2017knowledge}. Incorporating explicit knowledge allows for models to better recall factual information about specific entities \cite{knowbert}. Despite these successes, a persistent challenge that recent works continue to identify is how to leverage knowledge for low-resource regimes, such as \textit{tail} examples that appear rarely (if at all) in the training data \cite{fevry2020empirical}.

In this work, we study knowledge incorporation in the context of named entity disambiguation (NED) to better disambiguate the long tail of entities that occur infrequently during training.\footnote{In this work, we define tail entities as those occurring 10 or fewer times in the training data.} Humans disambiguate by leveraging subtle reasoning over entity-based knowledge to map strings to entities in a knowledge base. For example, in the sentence \emph{``Where is Lincoln in Logan County?''}, resolving the mention ``Lincoln'' to ``Lincoln, IL'' requires reasoning about {\em relations} because ``Lincoln, IL''---not ``Lincoln, NE'' or ``Abraham Lincoln''---is the capital of Logan County. Previous NED systems disambiguate by memorizing co-occurrences between entities and textual context in a self-supervised manner \cite{fevry2020empirical, yamada2019pre}. The self-supervision is critical to building a model that is easy to maintain and does not require expensive hand-curated features. However, these approaches struggle to handle tail entities: a baseline SotA model from \cite{fevry2020empirical} achieves less than \yell{28} F1 points over the tail, compared to \yell{86} F1 points over all entities. 

\begin{figure*}[t]
    \centering
    \includegraphics[width=0.98\linewidth]{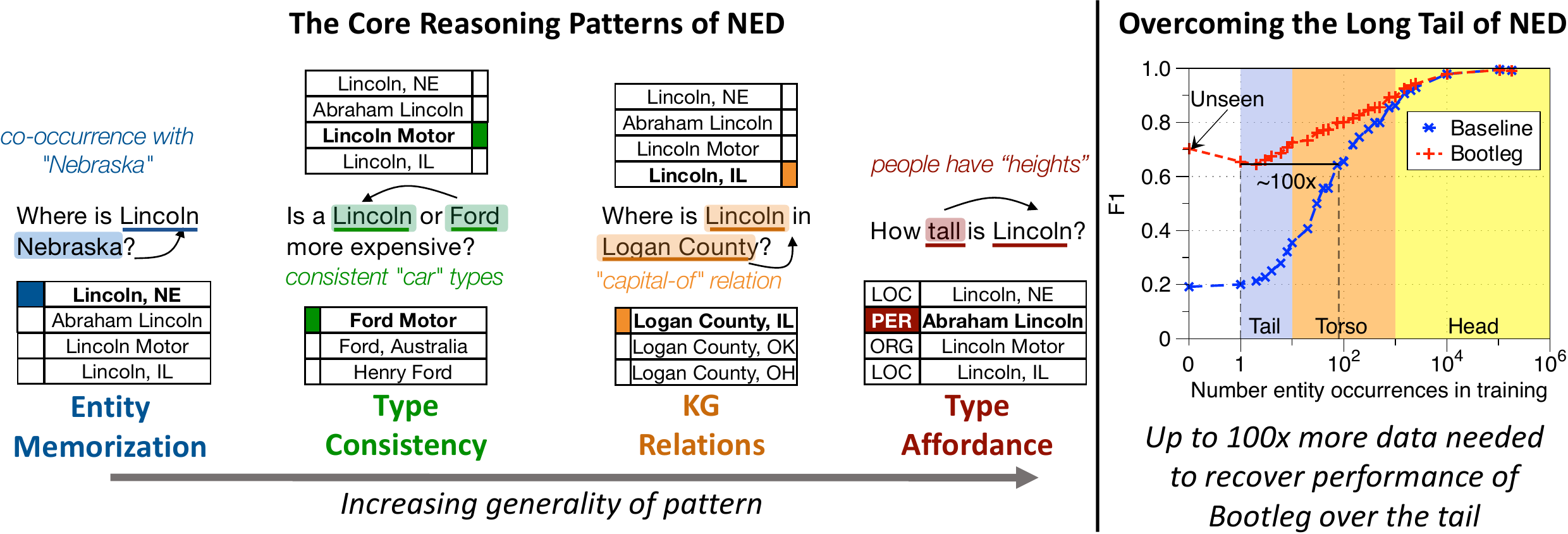}
    \caption{(Left) shows the four reasoning patterns for disambiguation. The correct entity is bolded. (Right) shows F1 versus number of times an entity was seen in training data for a baseline NED model compared to \name across the head, torso, tail, and unseen.}
    \label{fig:dataflow}
\end{figure*}

Despite their rarity in training data, many real-world entities are tail entities: 89\% of entities in the Wikidata knowledge base do not have Wikipedia pages to serve as a source of textual training data. However, to achieve 60 F1 points on disambiguation, we find that the prior SotA baseline model should see an entity on-the-order-of 100 times during training (\autoref{fig:dataflow} (right)). This presents a scalability challenge as there are 15x more entities in Wikidata than in Wikipedia, the majority of which are tail entities. For the model to observe each of these tail entities 100x, the training data would need to be scaled by 1,500x the size of Wikipedia. Prior approaches struggle with the tail, yet industry applications such as search and voice assistants are known to be tail-heavy \cite{bernstein2012direct, googlequeries}. Given the requirement for high quality tail disambiguation, major technology companies continue to press on this challenge \cite{nq2019google, petroni2020kilt}.

Instead of scaling the training data until co-occurrences between tail entities and text can be memorized, we define a principled set of reasoning patterns for entity disambiguation across the head and tail. When humans disambiguate entities, they leverage signals from context as well as from entity {\em relations} and {\em types}. 
For example, resolving ``Lincoln'' in the text \emph{``How tall is Lincoln?''} to ``Abraham Lincoln'' requires reasoning that people, not locations or car companies, have heights---a type affordance pattern. These core patterns apply to both head and tail examples with high coverage and involve reasoning over entity facts, relations, and types, information which is available for both head and tail in structured data sources. 
\footnote{We find that type affordance patterns apply to over \yell{84\%} of all examples, including tail examples, while KG relation patterns apply to over \yell{27\%} of all examples and type consistency applies to over \yell{8\%} of all examples. In Wikidata, \yell{75\%} of entities that are {\em not} in Wikipedia have type or knowledge graph connectivity signals, and among tail entities, \yell{88\%} are in non-tail type categories and \yell{90\%} are in non-tail relation categories.}
Thus, we hypothesize that these patterns assembled from the structured resources can be learned over training data and generalize to the tail.



In this work, we introduce \name, an open-source, self-supervised NED system designed to succeed on head \textit{and} tail entities. \footnote{\name is open-source at \url{http://hazyresearch.stanford.edu/bootleg}}
\name encodes the entity, relation, and type signals as embedding inputs to a simple stacked Transformer architecture. The key challenges we face are understanding how to use knowledge for NED, designing a model that learns those patterns, and fully extracting the useful knowledge signals from the training data:
\begin{itemize}
    \item \textbf{Tail Reasoning:} Humans use subtle reasoning patterns to disambiguate different entities, especially unfamiliar tail entities. The first challenge is characterizing these reasoning patterns and understanding their coverage over the tail. 
    \item \textbf{Poor Tail Generalization:} We find that a model trained using standard regularization and a combination of entity, type and relation information performs 10 F1 points {\em worse} on disambiguating unseen entities compared to the two models which respectively use only type and only relation information. We find this performance drop is due to the model's over-reliance on discriminative textual and entity features compared to more general type and relation features.
    \item \textbf{Underutilized Data:} Self-supervised models improve with more training data \cite{gpt3}. However, only a limited portion of the standard NED training dataset, Wikipedia, is useful: Wikipedia lacks labels~\cite{ghaddar2017winer} and we find that an estimated \yell{68\%} of entities in the dataset are not labeled.\footnote{We computed this statistic by computing the number of proper nouns and the number of pronouns/known aliases for an entity on that entity's page that were not already linked.}
\end{itemize}

\name addresses these challenges through three contributions: 

\begin{itemize}
    \item \textbf{Reasoning Patterns for Disambiguation:} We contribute a principled set of core disambiguation patterns for NED (\autoref{fig:dataflow} (left))---entity memorization, type consistency, KG relation, and type affordance---and show that on slices of Wikipedia examples exemplifying each pattern, \name provides a lift  over the baseline SotA model on tail examples by 18 F1, 56 F1, 62 F1, and 45 F1 points respectively. Overall, using these patterns, \name meets or exceeds state-of-the-art performance on three NED benchmarks and outperforms the prior SotA by more than 40 F1 points on the tail of Wikipedia. 
    \item \textbf{Generalizing Learning to the Tail:} Our key insight is that there are distinct entity-, type-, and relation- tails. Over tail entities (based on entity count in the training data), 88\% have non-tail types and 90\% have non-tail relations. The model should balance these signals differently depending on the particular entity being disambiguated. We thus contribute a new 2D regularization scheme to combine the entity, tail, and relation signals and achieve a lift of \yell{13.6} F1 points on unseen entities compared to the model using standard regularization techniques. We conduct extensive ablation studies to verify the effectiveness of our approach. 
    \item \textbf{Weak Labelling of Data:} 
    Our insight is that because Wikipedia is highly structured---most sentences on an entity’s Wikipedia page refer to that entity via pronouns or alternative names---we can weakly label our training data to label mentions. Through weak labeling, we increase the number of labeled mentions in the training data by 1.7x, and find this provides a \yell{2.6} F1 point lift on unseen entities.
\end{itemize}
With these three contributions, \name achieves SotA on three NED benchmarks. We further show that embeddings from \name are useful for downstream applications that require the knowledge of entities. We show the reasoning patterns learned in \name transfer to tasks beyond NED by extracting \name's learned embeddings and using them to set a new SotA by \yell{1.0} F1 points on the TACRED relation extraction task \cite{zhang2017tacred, Alt2020TACREDRA}, where the prior SotA model also uses entity-based knowledge \cite{knowbert}. \name representations further provide an 8\% performance lift on highly optimized industrial search and assistant tasks at a major technology company. For \name's embeddings to be viable for production, it is critical that these models are space-efficient: the models using only \name relation and type embeddings each achieve \yell{3.3x} the performance of the prior SotA baseline over unseen entities using \yell{1\%} of the space.

\section{NED Overview and Reasoning Patterns}
\label{sec:ned_primer}
We now define the task of named entity disambiguation (NED), the four core reasoning patterns, and the structural resources required for learning the patterns.

\para{Task Definition}
Given a knowledge base of entities $\mathcal{E}$ and an input sentence, the goal of named entity disambiguation is to determine the entities $e \in \mathcal{E}$ referenced in each sentence. Specifically, the input is a sequence of $N$ tokens $\mathcal{W} = \{w_1, \ldots, w_N\}$ and a set of $M$ non-overlapping spans in the sequence $\mathcal{W}$, termed \textit{mentions}, to be disambiguated $\mathcal{M} = \{m_1, \ldots, m_M\}$. The output is the most likely entity for each mention.

\paragraph{The Tail of NED}
We define the tail, torso, and head of NED as entities occurring less than 11 times, between 11 and 1,000, and more than 1,000 times in training, respectively. Following \autoref{fig:dataflow} (right), the head represents those entities a simple language-based baseline model can easily resolve, as shown by a baseline SotA model from \cite{fevry2020empirical} achieving \yell{86} F1 over all entities. These entities were seen enough times during training to memorize distinguishing contextual cues. The tail represents the entities these models struggle to resolve due to their rarity in training data, as shown by the same baseline model achieving less than \yell{28} F1 on the tail.

\subsection{Four Reasoning Patterns}
When humans disambiguate entities in text, they conceptually leverage signals over entities, relationships, and types. Our empirical analysis reveals a set of desirable reasoning patterns for NED. 
The patterns operate at different levels of granularity (see \autoref{fig:dataflow} (left))---from patterns which are highly specific to an entity, to patterns which apply to categories of entities---and are defined as follows. 
\begin{itemize}
    \item{\textbf{Entity Memorization:}} We define entity memorization as the factual knowledge associated with a specific entity. Disambiguating ``Lincoln'' in the text \emph{``Where is Lincoln, Nebraska?''} requires memorizing that ``Lincoln, Nebraska'', not ``Abraham Lincoln'' frequently occurs with the text ``Nebraska'' (\autoref{fig:dataflow} (left)). This pattern is easily learned by now-standard Transformer-based language models. As this pattern is at the {\em entity-level}, it is the least general pattern.
    
    
    \item{\textbf{Type Consistency:}} Type consistency is the pattern that certain textual signals in text indicate that the types of entities in a collection are likely similar. For example,  when disambiguating ``Lincoln'' in the text \emph{``Is a Lincoln or Ford more expensive?''}, the keyword ``or'' indicates that the entities in the pair (or sequence) are likely of the same Wikidata type, ``car company''. Type consistency is a more general pattern than entity memorization, covering \yell{12\%} of the tail examples in a sample of Wikipedia.\footnote{Coverage numbers are calculated from representative slices of Wikidata that require each reasoning pattern. Additional details in \autoref{sec:analysis}.}

    \item {\textbf{KG Relations:}} We define the knowledge graph (KG) relation pattern as when two candidates have a known KG relationship and textual signals indicate that the relation is discussed in the sentence. For example, when disambiguating ``Lincoln'' in the sentence \emph{``Where is Lincoln in Logan County?''}, ``Lincoln, IL'' has the KG relationship ``capital of'' with Logan County while Lincoln, NE does not. The keyword ``in'' is associated with the relation ``capital of'' between two location entities, indicating that ``Lincoln, IL'' is correct, despite being the less popular candidate entity associated with ``Lincoln''.  As patterns over pairs of entities with KG relations cover \yell{23\%} of the tail examples, this is a more general reasoning pattern than consistency.

    \item{\textbf{Type Affordance:}} We define type affordance as the textual signals associated with a specific entity-type in natural language. For example, ``Manhattan'' is likely resolved to the cocktail rather than the burrough in the sentence \emph{``He ordered a Manhattan.''} due to the affordance that drinks, not locations, are ``ordered''. As affordance signals cover \yell{76\%} of the tail examples, it is the most general reasoning pattern.

\end{itemize}

\paragraph{Required Structural Resources}
An NED system requires entity, relation, and type knowledge signals to learn these reasoning patterns. Entity knowledge is captured in unstructured text,
while relation signals and type signals are readily available in structured knowledge bases such as Wikidata: from a sample of Wikipedia, \yell{27}\% of all mentions and \yell{23}\% of tail mentions participate in a relation, and \yell{97}\% of all mentions and \yell{92}\% of tail mentions are assigned some type in Wikidata. As these structural resources are readily available for all entities, they are useful for generalizing to the tail. A rare entity with a particular type or relation can leverage textual patterns learned from every other entity with that type or relation.

Given the input signals and reasoning patterns, the next key challenge is ensuring that the model combines the discriminative entity and more general relation and type signals that are useful for disambiguation.

\section{\name Architecture for Tail Disambiguation}
\label{sec:architecture}

We now describe our approach to leverage the reasoning patterns based on entity, relation, and type signals. We then present our new regularization scheme to inject inductive bias of when to use general versus discriminative reasoning patterns and our weak labeling technique to extract more signal from the self-supervision training data. 


\subsection{Encoding the Signals}
We first encode the structural signals---entities, KG relations and types---by mapping each to a set of embeddings. 
\begin{itemize}[leftmargin=*, itemsep=3pt]
\item{\textit{Entity Embedding:}} Each entity $e$ is represented by a unique embedding $\mathbf{u}_e$.

\item{\textit{Type Embedding:}}
Let $\mathcal{T}$ be the set of possible entity types. Given a known mapping from an entity $e$ to its set $\{t_{e,1}, \ldots, t_{e,T} | t_{e,i} \in \mathcal{T}\}$ of $T$ possible types, \name assigns an embedding $\mathbf{t}_{e,i}$ to each type. 
Because an entity can have multiple types, we use an additive attention~\cite{bahdanau2014neural}, $\mathrm{AddAttn}$, to create a single type embedding $\mathbf{t}_e = \mathrm{AddAttn}([\mathbf{t}_{e,1}, \ldots, \mathbf{t}_{e,T}])$. 
We further allow the model to leverage coarse named entity recognition types through a mention-type prediction module (see \autoref{sec:appendix:model} for details). This coarse predicted type is concatenated with the assigned type to form $\mathbf{t}_e$. 

\item{\textit{Relation Embedding:}} Let $\mathcal{R}$ represent the set of possible relationships any entity can participate in. Similar to types, given a mapping from an entity $e$ to its set $\{r_{e,1}, \ldots, r_{e,R} | r_{e,i} \in \mathcal{R}\}$ of $R$ relationships, \name assigns an embedding $\mathbf{r}_{e,i}$ to each relation. Because an entity can participate in multiple relations, we use the additive attention to compute $\mathbf{r}_e = \mathrm{AddAttn}([\mathbf{r}_{e,1}, \ldots, \mathbf{r}_{e,R}])$.
\end{itemize}

As in existing work \cite{fevry2020empirical, Phan2019PairLinkingFC},
given the input sentence of length $N$ and set of $M$ mentions, \name generates for each mention $m_i$ a set $\Gamma(m_i) = \{e_i^1, \ldots, e_i^K\}$ of $K$ possible entity candidates that could be referred to by $m_i$.
For each candidate and its associated types and relations, \name uses a multi-layer perceptron $\mathbf{e} = \textrm{MLP}([\mathbf{u}_e, \mathbf{t}_e, \mathbf{r}_e])$
to generate a vector representation for each candidate entity, for each mention. We denote this entity matrix as $\mathbf{E} \in \mathbbm{R}^{M \times K \times H}$, where $H$ is the hidden dimension. We use BERT to generate contextual embeddings for each token in the input sentence. We denote this sentence embedding as $\mathbf{W} \in \mathbbm{R}^{N \times H}$. $\mathbf{W}$ and $\mathbf{E}$ are passed to \name's model architecture, described next.

\begin{figure}
\centering
  \includegraphics[width=.5\linewidth]{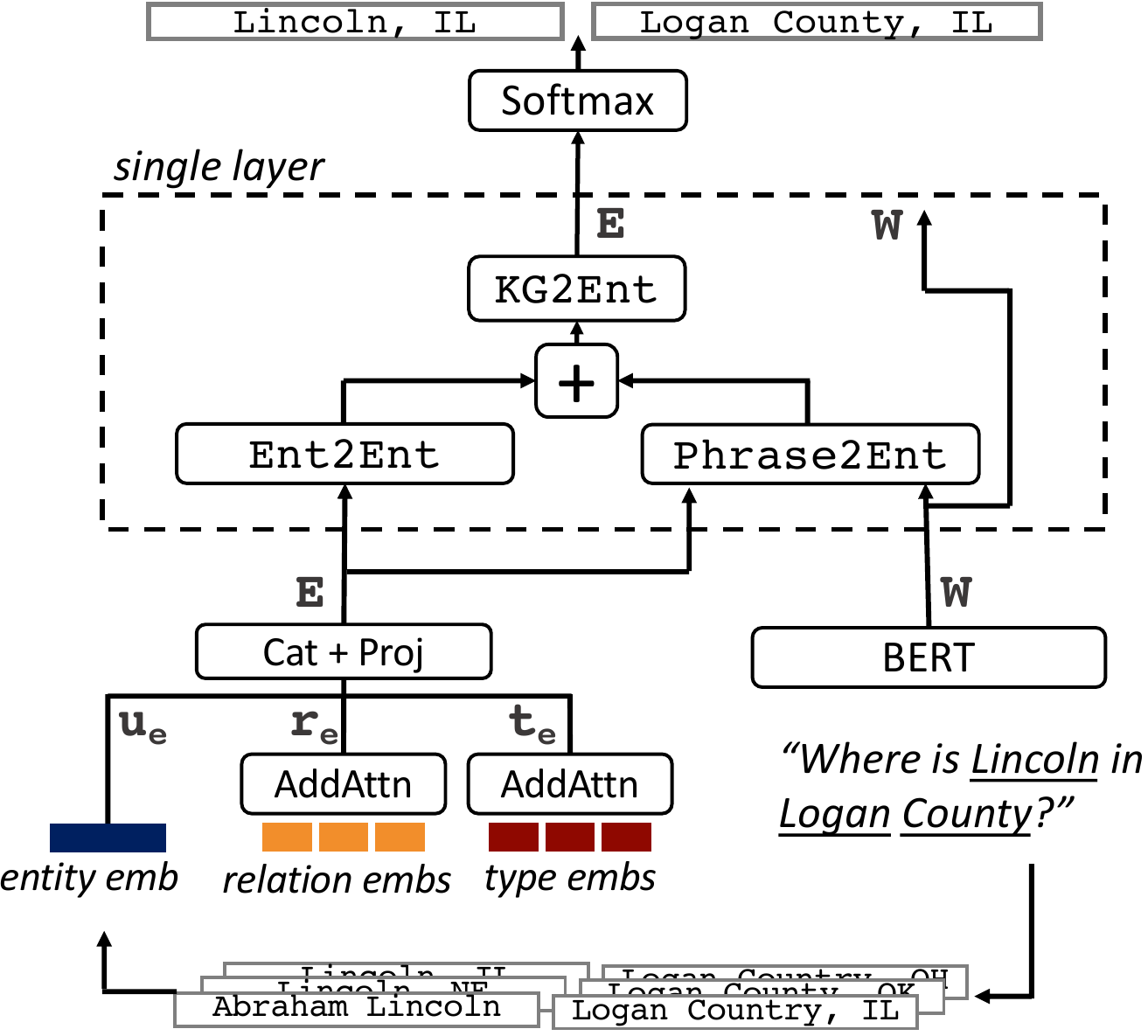}
  \captionof{figure}{\name's neural model. The entity, type, and relation embeddings are generated for each candidate and concatenated to form our entity representation matrix $\mathbf{E}$. This, together with our word embedding matrix $\mathbf{W}$, are inputs to \name's \texttt{Ent2Ent}, \texttt{Phrase2Ent}, and \texttt{KG2Ent} modules which aim to encode the four reasoning patterns. The most likely candidate for each mention is returned.}
  \label{fig:test1}
\end{figure}


\subsection{\name Model Architecture}
\label{sec:architecture:model}

The design goal of \name is to capture the reasoning patterns by modeling textual signals associated with entities (for entity memorization), co-occurrences between entity types (for type consistency), textual signals associated with relations along with which entities are explicitly linked in the KG (for KG relations), and textual signals associated with types (for type affordance). We design three modules to capture these design goals: a phrase memorization module, a co-occurrence memorization module, and a knowledge graph connection module. The model architecture is shown in \autoref{fig:test1}. We describe each module next.




\paragraph{Phrase Memorization Module} We design the phrase memorization module, \texttt{Phrase2Ent}, to encode the dependencies between the input text and the entity, relation, and type embeddings. The purpose of this module is to learn textual cues for the entity memorization and type affordance patterns. It should also learn relation context for the KG relation pattern. It will, for example, allow the person type embedding to encode the association with the keyword ``height''. The module accepts as input $\mathbf{E}$ and $\mathbf{W}$ and outputs 
$\mathbf{E}_p = \textrm{MHA}(\mathbf{E}, \mathbf{W})$, 
where MHA is the standard multi-headed attention with a feed-forward layer and skip connections \cite{vaswani2017attention}. 

\paragraph{Co-occurrence Memorization Module} We design the co-occurrence memorization module, \texttt{Ent2Ent}, to encode the dependencies between entities. The purpose of the \texttt{Ent2Ent} module is to learn textual cues for the type consistency pattern. The module accepts $\mathbf{E}$ and computes $\mathbf{E}_c = \textrm{MHA}(\mathbf{E})$ using self-attention.

\paragraph{Knowledge Graph (KG) Connection Module} We design the KG module, \texttt{KG2Ent}, to collectively resolve entities based on pairwise connectivity features. Let $\mathbf{K}$ represent the adjacency matrix of a (possibly weighted) graph where the nodes are entities and an edge between $e_i$ and $e_j$ signifies that the two entities share some pairwise feature. Given $\mathbf{E}$, \texttt{KG2Ent} computes
$\mathbf{E}_k = \textrm{softmax}(\mathbf{K} + w\mathbf{I})\mathbf{E} + \mathbf{E}$
where $\mathbf{I}$ is the identity and $w$ is a learned scalar weight that allows \name to learn to balance the original entity and its connections.  This module allows for representation transfer between two related entities, meaning entities with a high-scoring representation will boost the score of related entities. The second computation acts as a skip connection between the input and output. In \name, we allow the user to specify multiple \texttt{KG2Ent} modules: one for each adjacency matrix. The purpose of \texttt{KG2Ent} along with \texttt{Phrase2Ent} is to learn the KG relation pattern.

\paragraph{End-to-End}
The computations for one layer of \name includes:
\begin{align*}
    \mathbf{E}' =& \textrm{MHA}(\mathbf{E}, \mathbf{W}) + \textrm{MHA}(\mathbf{E}) \\
    \mathbf{E}_k =& \textrm{softmax}(\mathbf{K} + w\mathbf{I})\mathbf{E}' + \mathbf{E}'
\end{align*}
where $\mathbf{E}_k$ is passed as the entity matrix to the next layer. After the final layer, \name scores each entity by computing
$\mathbf{S}_{dis} = \max(\mathbf{E}_k \mathbf{v}^T, \mathbf{E}' \mathbf{v}^T)$
with $\mathbf{S}_{dis} \in \mathbbm{R}^{M \times K}$ and learned scoring vector $\mathbf{v} \in \mathbbm{R}^{H}$. \name then outputs the highest scoring candidate for each mention. This scoring treats $\mathbf{E}_k$ and $\mathbf{E}'$ as two separate predictions in an ensemble method, allowing the model to use collective reasoning from $\mathbf{E}_k$ when it achieves the highest scoring representation. If there are multiple \texttt{KG2Ent} modules, we use the average of their outputs as input to the next layer and, for scoring, take the maximum score across all outputs. For training, we use the cross-entropy loss of $\mathbf{S}$ to compute the disambiguation loss $\mathcal{L}_{dis}$.

\subsection{Improving Tail Generalization}
\label{sec:architecture:tail_gen}
Regularization is the standard technique to encourage models to generalize, as models will naturally fit to discriminative features. However, we demonstrate that standard regularization is not effective when we want to leverage a combination of general and discriminative signals. We then present two techniques, regularization and weak labeling, to encourage \name to incorporate general structural signals and learn general reasoning patterns.

\subsubsection{Regularization}
We hypothesize that \name will over-rely on the more discriminative entity features compared to the more general type and relation features to lower training loss. However, tail disambiguation requires \name to leverage the general features.
Using standard regularization techniques, we evaluate three models which respectively use only type embeddings, only relation embeddings, and a combination of type, relation, and entity embeddings. \name's performance on unseen entities is \yell{10} F1 points worse on the latter than each of the former two, suggesting that standard regularization is not sufficient when the signals operate at different granularities (details \autoref{tab:micro_ablation} in \autoref{sec:appendix:results}). 

We can improve tail performance if \name leverages memorized discriminative features for popular entities and general features for rare entities. We achieve this by designing a new regularization scheme for the entity-specific embedding $\mathbf{u}$, which has two key properties: it is 2-dimensional and more popular entities are regularized less than less popular ones. 
\begin{itemize}
    \item \textit{2-dimensional}: In contrast to 1-dimensional dropout, 2-dimensional regularization involves masking the full embedding. With probability $p(e)$, we set $\mathbf{u} = \mathbf{0}$ {\em before} the MLP layer; i.e., $\mathbf{e} = \textrm{MLP}([\mathbf{0}, \mathbf{t}_e, \mathbf{r}_e])$. Entirely masking the entity embedding in these cases, the model learns to disambiguate using the type and relation patterns, without entity knowledge.
    \item \textit{Inverse Popularity}: We find in ablations (\autoref{sec:appendix:results}) that setting $p(e)$ proportional to the power of the inverse of the entity $e$'s popularity in the training data (i.e., the more popular the less regularized), gives us the best performance and improves by \yell{13.6 F1} on unseen entities over standard regularization. In contrast, fixing $p(e)$ at $80$\% improves performance by over \yell{11.3} F1 over standard regularization, and regularizing proportional to the power of popularity only improves performance by \yell{3.8} F1 (details in \autoref{sec:evaluation}).
\end{itemize}
The regularization scheme encourages \name to use entity-specific knowledge when the entity is seen enough times to memorize entity patterns and encourages the use of generalizable patterns over the rare, highly masked, entities.


\subsubsection{Weakly Supervised Data Labeling}
We use Wikipedia to train \name: we define a self-supervision task in which the internal links in Wikipedia are the gold entity labels for mentions during training. Although this dataset is large and widely used, it is often incomplete with an estimated \yell{68\%} of named entities being unlabeled. Given the scale and the requirement that \name be self-supervised, it is not feasible to hand-label the data. Our insight is that because Wikipedia is highly structured---most sentences on an entity's Wikipedia page refer to that entity via pronouns or alternative names---we can {\em weakly label} our training data \cite{ratner2017snorkel} to label mentions. We use two heuristics for weak labeling: the first labels pronouns that match the gender of a person's Wikipedia page as references to that person, and the second labels known alternative names for an entity if the alternative name appears in sentences on the entity's Wikipedia page. Through weak labeling, we increase the number of labeled mentions in the training data by \yell{1.7x} across Wikipedia, and find this provides a \yell{2.6 F1} lift on unseen entities (full results in \autoref{sec:appendix:results} \autoref{tab:micro_ablation_wl}).


\section{Experiments}
\label{sec:evaluation}

We demonstrate that \name (1) nearly matches or exceeds state-of-the-art performance on three standard NED benchmarks and (2) outperforms a BERT-based NED baseline on the tail. As NED is critical for downstream tasks that require the knowledge of entities, we (3) verify \name's learned reasoning patterns can transfer by using them for a downstream task: using \name's learned representations, we achieve a new SotA on the TACRED relation extraction task and improve performance on a production task at a major technology company by 8\%. Finally, we (4) demonstrate that \name can be sample-efficient by using only a fraction of its learned entity embeddings without sacrificing performance. We (5) ablate \name to understand the impact of the structural signals and the regularization scheme on improved tail performance.  

\subsection{Experimental Setup}

\paragraph{Wikipedia Data}
We define our knowledge base as the set of entities with mentions in Wikipedia (for a total of \yell{5.3M} entities). We allow each mention to have up to $K=30$ possible candidates. As \name is a sentence disambiguation system, we train on individual sentences from Wikipedia, where the anchor links and our weak labeling (\autoref{sec:architecture:tail_gen}) serve as mention labels.

Our candidate lists $\Gamma$ are mined from Wikipedia anchor links and the ``also known as'' field in Wikidata. For each person, we further add their first and last name as aliases linking to that person. We use the mention boundaries provided in the Wikipedia data and generate candidates by performing a direct lookup in $\Gamma$.

We use Wikidata and YAGO knowledge graphs and Wikipedia to extract structural data about entity types and relations as input for \name. Further details about data are in \autoref{sec:appendix:results}.

\paragraph{Metrics} We report micro-average F1 scores for all metrics over true anchor links in Wikipedia (not weak labels). We measure the torso and tail sets based on the number of times that an entity is the gold entity across Wikipedia anchors and weak labels, as this represents the number of times an entity is seen by \name. For benchmarks, we also report precision and recall using the number of mentions extracted by \name and the number of mentions defined in the data as denominators, respectively. The numerator is the number of correctly disambiguated mentions. For Wikipedia data experiments, we filter mentions such that (a) the gold entity is in the candidate set and (b) they have more than one possible candidate. The former is to decouple candidate generation from model performance for ablations.\footnote{We drop only 1\% of mentions from this filter.} The latter is to not inflate a model's performance, as all models are trivially correct when there is a single candidate. 

\paragraph{Training}
For our main \name model, we train for two epochs on Wikipedia sentences with a maximum sentence length of $100$. For our benchmark model, we train for one epoch and additionally add a title embedding feature, a sentence co-occurrence KG matrix as another KG module, and a Wikipedia page co-occurrence statistical feature. Additional details about the models and training procedure are in \autoref{sec:appendix:results}.

\begin{table}[t]
    \begin{center}
    \caption{We compare \name to the best published numbers on three NED benchmarks. ``-'' indicates that the metric was not reported. Bolded numbers indicate the best value 
    for each metric on each benchmark.
    }
    \vspace{3mm}
    \begin{tabular}{llccc}
    \toprule
    Benchmark               &  Model                        &      Precision &        Recall &             F1  \\
    \midrule
    \multirow{2}{*}{KORE50} & \citet{shengzeentity}\footnote{Although \citet{shengzeentity} does end-to-end entity linking, their reported KORE50 results is the current SotA, beating the result of 78 from (\citet{Phan2019PairLinkingFC})}         &         80.0  &         79.8  &         79.9  \\
                            & \name                         & \yell{\textbf{86.0}} & \yell{\textbf{85.4}} & \yell{\textbf{85.7}} \\
    \midrule
    \multirow{2}{*}{RSS500} & \citet{Phan2019PairLinkingFC} &         82.3  & 82.3 &        82.3  \\
                            & \name                         & \yell{\textbf{82.5}} & \yell{\textbf{82.5}} & \yell{\textbf{82.5}} \\

    \midrule
    \multirow{2}{*}{AIDA}   & \citet{fevry2020empirical}    &            -   & \textbf{96.7} &            -   \\
                            & \name                         & \yell{96.9} & \yell{\textbf{96.7}} & \yell{96.8} \\
    \bottomrule
    \end{tabular}
    \label{tab:benchmark-nums}
    \end{center}
\end{table}

\subsection{\name Performance}
\para{Benchmark Performance}  
To understand the overall performance of \name, we compare against reported state-of-the-art numbers of two standard sentence benchmarks (KORE50, RSS500) and the standard document benchmark (AIDA CoNLL-YAGO). Benchmark details are in \autoref{sec:appendix:results}.

For AIDA, we first convert each document into a set of sentences where a sentence is the document title, a BERT SEP token, and the sentence. We find this is sufficient to encode document context into \name. We fine-tune the pretrained \name model on the AIDA training set with learning rate of $0.00007$, $2$ epochs, batch size of $16$, and evaluating every $25$ steps. We choose the test score associated with the best validation score.\footnote{We use the standard candidate list from \citet{pershina} when comparing to existing systems for fine-tuning and inference for AIDA CoNLL-YAGO.} In Table~\ref{tab:benchmark-nums}, we show that \name achieves up to \yell{5.8} F1 points higher than prior reported numbers on benchmarks.

\begin{table}[ht]
\begin{center}
 \caption{(top) We compare \name to a BERT-based NED baseline (NED-Base) on validation sets of a Wikipedia dataset. We report micro-average F1 scores. All torso, tail, and unseen validation sets are filtered by the number of entity occurrences in the training data and such that the mention has more than one candidate.}
    \vspace{3mm}
    \label{tab:tail}
\begin{tabular}{lcccc}
\toprule
 Model           & All Entities & Torso Entities & Tail Entities & Unseen Entities \\
            \midrule

NED-Base    & 85.9         & 79.3           & 27.8          & 18.5             \\
\name     & \textbf{91.3}         & \textbf{87.3}           & \textbf{69.0}  & \textbf{68.5}            \\

\midrule
\name (Ent-only) & 85.8         & 79.0           & 37.9          & 14.9            \\
\name (Type-only)   & 88.0         & 81.6           & 62.9          & 61.6            \\
\name (KG-only)    &    87.1          &  79.4     &   64.0     &  64.7        \\
\midrule
\# Mentions & 4,065,778      & 1,911,590        & 162,761         & 9,626     
\\
\bottomrule
\end{tabular}
\end{center}
\end{table}

\para{Tail Performance} To validate that \name improves tail disambiguation, we compare against a baseline model from \citet{fevry2020empirical}, which we refer to as NED-Base.\footnote{As code for the model from \citet{fevry2020empirical} is not publicly available, we re-implemented the model. We used our candidate generators and fine-tuned a pretrained BERT encoder rather than training a BERT encoder from scratch, as is done in \citet{fevry2020empirical}. We trained NED-Base on the same weak labelled data as \name for 2 epochs.} NED-Base learns entity embeddings by maximizing the dot product between the entity candidates and fine-tuned BERT-contextual representations of the mention.

NED-Base is successful overall on the validation achieving \yell{85.9} F1 points, which is within \yell{5.4} F1 points of \name (Table~\ref{tab:tail}). However, when we examine performance over the torso and tail, we see that \name outperforms NED-Base by \yell{8} and \yell{41.2} F1 points, respectively. Finally, on unseen entities, \name outperforms NED-Base by \yell{50} F1 points. Note that NED-Base only has access to textual data, indicating that text is often sufficient for popular entities, but not for rare entities.

\subsection{Downstream Evaluation}
\para{Relation Extraction}
Using the learned representations from \name, we achieve the new state-of-the-art on TACRED, a standard relation extraction benchmark. TACRED involves identifying the relationship between a specified subject and object in an example sentence as one of 41 relation types (e.g., spouse) or no relation. Relation extraction is a well-suited for evaluating \name because the substrings in the text can refer to many different entities, and the disambiguated entities impact the set of likely relations.

Given an example, we run inference with the \name model to disambiguate named entities and generate the contextual \name entity embedding matrix, which we feed to a simple Transformer architecture that uses SpanBERT~\cite{spanbert} (details in \autoref{sec:appendix:downstream}). We achieve a micro-average test F1 score of 80.3, which improves upon the prior state of the art---KnowBERT~\cite{knowbert}, which also uses entity-based knowledge---by 1.0 F1 points and the baseline SpanBERT model by 2.3 F1 points on TACRED-Revisited data (Table~\ref{tab:tacred}) (\cite{zhang2017tacred}, \citet{Alt2020TACREDRA}). We find that the \name downstream model corrects errors made by the SpanBERT baseline, for example by leveraging entity, type, and relation information or recognizing that different textual aliases refer to the same entity (see \autoref{tab:tacred_examples}).

\begin{center}
\begin{table}[b]
   \centering
   \captionof{table}{Test micro-average F1 score on revised TACRED dataset.}
   \begin{tabular}{lcc}
    \toprule
    Validation Set       & F1  \\
    \midrule
    \name Model  &   \textbf{80.3} \\
    KnowBERT &  79.3   \\
    SpanBERT &      78.0 \\
    \bottomrule
    \end{tabular}
   \label{tab:tacred}
\end{table}
\end{center}

In studying the slices for which the \name downstream model improves upon the baseline SpanBERT model, we rank TACRED examples in three ways: by the proportion of words where \name disambiguates it as an entity, leverages Wikidata relations for the embedding, and leverages Wikidata types for the embedding. For each of these three, we report the gap between the SpanBERT model and Bootleg model's error rates on the examples with above-median proportion (more \name signal) relative to the below-median proportion (less \name signal). We find that the relative gap between the baseline and Bootleg error rates is larger on the slice above (with more \name information) than below the median by 1.10x, 4.67x, and 1.35x respectively: with more \name information, the improvement our SotA model provides over SpanBERT increases (more details in \autoref{sec:appendix:downstream}).


\begin{table}[]
\begin{center}
 \caption{The following are examples of how the contextual entity representation from \name, generated from entity, relation, and type signals, can help our downstream model. We provide the TACRED example, signals provided by \name, as well our model and the baseline SpanBERT models' predictions.}
    \vspace{3mm}
    \label{tab:tacred_examples}
\begin{tabular}{p{5.75cm} p{5.5cm} p{1.75cm} p{1.75cm}}
\toprule
 TACRED Example & \name Signals & Our \newline Prediction & SpanBERT Prediction\\
            \midrule

\emph{Vincent Astor, like Marshall (subj), died unexpectedly of a heart attack (obj) in 1959 …} \newline\newline \textbf{Gold relation: Cause of Death} &  Disambiguates ``Marshall'' to Thomas Riley Marshall and ``heart attack'' to myocardial infarction, which have the Wikidata relation ``cause of death'' & Cause of \newline Death & No \newline Relation\\
\midrule
 \emph{The International Water Management (obj) Institute or IWMI (subj) study said both ….}  \newline\newline \textbf{Gold relation: Alternate Names}  &  Disambiguates alias ``International Water Management Institute'' and its acronym, the alias ``IWMI'', to the same Wikidata entity
   &  Alternate Names  &  No \newline Relation  \\
\bottomrule
\end{tabular}
\end{center}
\end{table}

\paragraph{Industry Use Case}
We additionally demonstrate how the learned entity embeddings from \name provide useful information to a system at a large technology company that answers factoid queries such as \emph{``How tall is the president of the United States?"}. We use \name's embeddings in the Overton~\cite{R2020OvertonAD} system and compare to the same system without \name embeddings as the baseline. We measure the overall test quality (F1) on an in-house entity disambiguation task as well as the quality over the tail slices which include unseen entities. Per company policy, we report relative to the baseline rather than raw F1 score; for example, if the baseline F1 score is $80.0$ and the subject F1 is $88.0$, the relative quality is $88.0/80.0 = 1.1$. Table~\ref{tab:overton} shows that the use of \name's embeddings consistently results in a positive relative quality, even over Spanish, French, and German, where improvements are most visible over tail entities.

\begin{center}
\begin{table}[b]
   \centering
   \captionof{table}{Relative F1 quality of an Overton\cite{R2020OvertonAD} model with \name embeddings over one without in four languages.}
   \begin{tabular}{lcccc}
    \toprule
    Validation Set       & English & Spanish & French & German \\
    \midrule
    All Entities  &    1.08 &    1.03 &   1.02 &   1.00 \\
    Tail Entities &    1.08 &    1.17 &   1.05 &   1.03 \\
    \bottomrule
    \end{tabular}
   \label{tab:overton}
\end{table}
\end{center}

\subsection{Memory Usage}
We explore the memory usage of \name during inference and demonstrate that by only using the entity embeddings for the top 5\% of entities, ranked by popularity in the training data, \name reduces its embedding memory consumption by 95\%, while sacrificing only \yell{0.8 F1} points over all entities. We find that the 5.3M entity embeddings used in \name consume the most memory, taking \yell{5.2} GB of space while the attention network only consumes 39 MB (1.37B updated model parameters in total, 1.36B from embeddings). As \name's representations must be used in a variety of downstream tasks, the representations must be memory-efficient: we thus study the effect of reducing \name's memory footprint by only using the most popular entity embeddings.

\begin{figure}[t]
    \centering
    \includegraphics[width=0.4\linewidth]{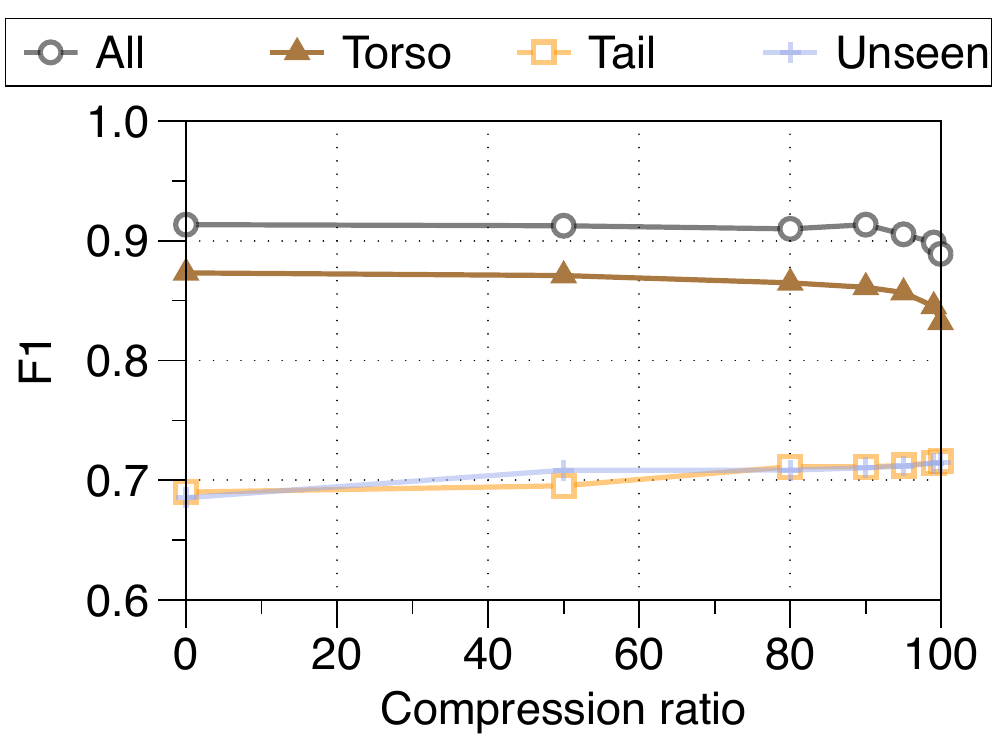}
    \caption{We show the error across all entities, torso entities, tail entities, and unseen entities as we decrease the number embeddings we use during inference, assigning the non-popular entities to a fixed unseen entity embedding. For example, a compression ratio of 80 means only the top 20\% of entity embeddings are used, ranked by entity popularity.}
    \label{fig:memory_eval}
\end{figure}

Specifically, for the top $k$\% of entities ranked by the number of occurrences in training data, we keep the learned entity embedding intact. For the remaining entities, we choose a random entity embedding for an unseen entity to use instead. 
Instead of storing 5.3M entity embeddings, we thus store $((100-k)/100)*5.3M$, which gives a compression ratio of $(100-k)$. \autoref{fig:memory_eval} shows performance for $k$ of 100, 50, 20, 10, 5, 1, and 0.1. We see that when just the top 5\% of entity embeddings are used, we only sacrifice $0.8$ F1 points overall and in fact score $2$ F1 points {\em higher} over the tail. We hypothesize that the increase in tail performance is due to the fact that the majority of mention candidates all have the same learned embedding, decreasing the amount of conflict among candidates from textual patterns.

\subsection{Ablation Study}
\paragraph{\name}
To better understand the performance gains of \name, we perform an ablation study over a subset of Wikipedia (data details explained in \autoref{sec:appendix:results}). We train \name with: (1) only learned entity embeddings (Ent-only), (2) only type information from type embeddings (Type-only), and (3) only knowledge graph information from relation embeddings and knowledge graph connections (KG-only). All model sizes are reported in \autoref{sec:appendix:results} \autoref{tab:model_sizes}. In Table~\ref{tab:tail}, we see that just using type or knowledge graph information leads to improvements on the tail of over \yell{25} F1 points and on the unseen entities of over \yell{46} F1 points compared to the Ent-only model. However, neither the Type-only nor KG-only model performs as well on any of the validation sets as the full \name model. An interesting comparison is between Ent-only and NED-Base. NED-Base overall outperforms Ent-only due to the fine-tuning of BERT word embeddings. We attribute the high performance of Ent-only on the tail compared to NED-Base to our \texttt{Ent2Ent} module which allows for memorizing co-occurrence patterns over entities.

\paragraph{Regularization}
\begin{table}[t]
\begin{center}
\captionof{table}{We show the micro F1 score over unseen entities for a Wikipedia sample as we vary the entity regularization scheme $p(e)$. A scalar percent means a fixed regularization. InvPop (inverse poularity scheme) applies less regularization for more popular entities and Pop applies more regularization for more popular entities.}
\begin{tabular}{lllllll}
\toprule
$p(e)$ &  $0\%$ &  $20\%$ &  $50\%$ & $80\%$ &  Pop &  InvPop \\
\midrule
Unseen Entities  &   48.6 &   52.5 &   57.7 &  59.9 & 52.4 & \textbf{62.2}   \\
\bottomrule
\end{tabular}
\label{tab:regularization}
\end{center}
\end{table}


To understand the impact of our entity regularization function $p(e)$ on overall performance, we perform an ablation study on a sample of Wikipedia (explained in \autoref{sec:appendix:results}). We apply (1) a fixed regularization set to a constant percent of 0, 20, 50 and 80, (2) a regularization function proportional to the power of the inverse popularity, and (3) the inverse of (2). \autoref{tab:regularization} shows results over unseen entities (full results and details in \autoref{sec:appendix:results}). We see that the fixed regularization of $80\%$ achieves the highest F1 over the fixed regularizations of (1). The method that regularizes by inverse popularity achieves the highest overall F1. We further see that the scheme where popular entities are more regularized sees a drop of \yell{9.8 F1} points in performance compared to the inverse popularity scheme.

\section{Analysis}
\label{sec:analysis}

We have shown that \name excels on benchmark tasks and that \name's learned patterns can transfer to non-NED tasks. We now verify whether the defined entity, type consistency, KG relation, and affordance reasoning patterns are responsible for these results. We evaluate each over a representative slice of the Wikipedia validation set that exemplifies one of the reasoning patterns and present the results from each ablated model (\autoref{tab:tail_analysis}). 

\begin{itemize}
    \item \textbf{Entity} To evaluate whether \name captures factual knowledge about entities in the form of textual entity cues, we consider the slice of \yell{28K} overall, \yell{5K} tail examples where the gold entity has no relation or type signals available. 
    \item \textbf{Type Consistency} To evaluate whether \name captures consistency patterns, we consider the slice of \yell{312K} overall, \yell{19K} tail examples that contain a list of three or more sequential distinct gold entities, where all items in the list share at least one type. 
    \item \textbf{KG Relation} To evaluate whether \name captures KG relation patterns, we consider the slice of \yell{1.1M} overall, \yell{37K} tail examples for which the gold entities are connected by a known relation in the Wikidata knowledge graph. 
    \item \textbf{Type Affordance} To evaluate whether \name captures affordance patterns, we consider a slice where the sentence contains keywords that are afforded by the type of the gold entity. We mine the keywords afforded by a type by taking the \yell{15} keywords that receive the highest TF-IDF scores over training examples with that type. This slice has \yell{3.4M} overall, \yell{124K} tail examples.
\end{itemize}

\paragraph{Pattern Analysis} For the slice representing each reasoning pattern, we find that \name provides a lift over the Entity-only and NED-Base models, especially over the tail. We find that \name generally combines the entity, relation, and type signals effectively, performing better than the individual Entity-only, KG-only, and Type-only models, although the KG-only model performs well on the KG relation slice. The lift from \name across slices indicates the model's ability to capture the reasoning required for the slice. We provide additional details in \autoref{sec:appendix:error}.

\begin{table}[]
\begin{center}
 \caption{We report the \textit{Overall/Tail} F1 scores across each ablation model for a slice of data that exemplifies a reasoning pattern. Each slice is representative but may not cover every example that contains the reasoning pattern.}
    \vspace{3mm}
    \label{tab:tail_analysis}
\begin{tabularx}{0.86\textwidth}{lcccc}
\toprule
 Model           & Entity & Type Consistency & KG Relation & Type Affordance \\
            \midrule

NED-Base    & 59/29         & 84/29   &   91/30  & 87/28             \\
\name     & \textbf{66/47}  & \textbf{95/85} & \textbf{98}/92  & \textbf{93/73}            \\

\midrule
\name (Ent-only) & 59/31    & 87/45   &  90/42  & 87/39            \\
\name (Type-only)   & 53/44     &  93/80    &  93/69  & 90/66            \\
\name (KG-only)    &    40/29  &  92/79   &   97/\textbf{93}   &  89/68        \\
\midrule
\% Coverage & 0.7\%/3.3\%   & 8\%/12\%   & 27\%/23\%   & 84\%/76\%     
\\
\bottomrule
\end{tabularx}
\end{center}
\end{table}

\paragraph{Error Analysis} We next study the errors made by \name and find four key error buckets. 
\begin{itemize}
    \item \textbf{Granularity} \name struggles with \textit{granularity}, predicting an entity that is too general or too specific compared to the gold entity (example in \autoref{tab:error_examples}). Considering the set of examples where the predicted entity is a Wikidata \textit{subclass} of the gold entity or vice versa, \name predicts a too general or specific entity in 12\% of overall and 7\% of tail errors.
    \item \textbf{Numerical} \name struggles with entities containing numerical tokens, which may be due to the fact that the BERT model represents some numbers with sub-word tokens and is known to not perform as well for numbers as other language models~\cite{wallace2019nlp} (example in \autoref{tab:error_examples}). To evaluate examples requiring reasoning over numbers, we consider the slice of data where the entity title contains a year, as this is the most common numerical feature in a title. This slice covers 14\% of overall and 25\% of tail errors.
    \item \textbf{Multi-Hop} There is room for improvement in multi-hop reasoning. In the example shown \autoref{tab:error_examples}, none of the present gold entities---Stillwater Santa Fe Depot, Citizens Bank Building (Stillwater, Oklahoma), Hoke Building (Stillwater, Oklahoma), or Walker Building (Stillwater, Oklahoma)---are directly connected in Wikidata; however, they share connections to the entity ``Oklahoma''. This indicates that the correct disambiguation is Citizens Bank Building (Stillwater, Oklahoma), not Citizens Bank Building (Burnsville, North Carolina). To evaluate examples requiring 2-hop reasoning, we consider examples where none of the present entities are directly linked in the KG, but a present pair connects to a different entity that is not present in the sentence. We find this occurs in 6\% of overall and 7\% of tail errors. This type of error represents a fundamental limitation of \name as we do not encode any form of multi-hop reasoning over a KG in \name. Our KG information only encodes single-hop patterns (i.e., direct connections).
    \item \textbf{Exact Match} \name struggles on several examples in which the exact entity title is present in the text. Considering examples where the BERT Baseline is correct but \name is incorrect, in 28\% of the examples, the textual mention is an exact match of the entity title. Further, 32\% of the examples contain a keyword from the entity title that \name misses (example in \autoref{tab:error_examples}).  We attribute this decrease in performance to \name's regularization. This mention-to-entity similarity would need to be encoded in \name's entity embedding, but the regularization encourages \name to not use entity-level information.
\end{itemize} 

\begin{table}[]
\begin{center}
 \caption{We identify four key error buckets for \name: granularity, numerical errors, multi-hop reasoning, and missed exact matches. We provide a Wikipedia example, the gold entity, and \name's predicted entity for each example.}
    \vspace{3mm}
    \label{tab:error_examples}
\begin{tabular}{p{1.75cm} p{5.5cm} p{3.4cm} p{3.7cm}}
\toprule
 Error  & Wikipedia Example & \name Prediction & Gold Entity\\
            \midrule

Granularity    &\emph{Posey is the recipient of a Golden Globe Award nomination, a \textbf{Satellite Award} nomination and two Independent Spirit Award nominations.} &    Satellite Awards      & Satellite Award for Best Actress – Motion Picture \\
\midrule
Numerical   &   \emph{He competed in the individual road race and team time trial events at the \textbf{1976 Summer Olympics}.}    &  Cycling at the 1960 Summer Olympics – 1960 Men’s Road Race    &    Cycling at the 1976 Summer Olympics – 1976 Men’s Road Race      \\
\midrule
Multi-hop  &  \emph{Other nearby historic buildings include the Santa Fe Depot, the \textbf{Citizens Bank Building}, the Hoke Building, the Walker Building, and the Courthouse}   & Citizens Bank Building (Burnsville, North Carolina) &  Citizens Bank Building (Stillwater, Oklahoma)     \\
\midrule
Exact Match  &   \emph{According to the \textbf{Nielsen Media Research}, the episode was watched by 469 million viewers...}  & Nielsen ratings &  Nielsen Media Research 
\\
\bottomrule
\end{tabular}
\end{center}
\end{table}


\section{Related Work}
\label{sec:related_work}
We discuss related work in terms of both NED and the broader picture of self-supervised models and tail data. Standard, pre-deep-learning approaches to NED have been rule-based~\cite{aberdeen1996mitre} or leverage statistical techniques and manual feature engineering to filter and rank candidates~\cite{yadav2019survey}. For example, link counts and similarity scores between entity titles and mention are two such features~\cite{cucerzan2007large}. These systems tend to be hard to maintain over time, with the work of \citet{petasis2001using} building a model to detect when a rule-based NED system needs to be retrained and updated.

In recent years, deep learning systems have become the new standard (see \citet{mudgal2018deep} for a high-level overview of deep learning approaches to entity disambiguation and entity matching problems). The most recent state-of-the-art models generally rely on deep contextual word embeddings with entity embeddings \cite{fevry2020empirical, yamada2019pre, shahbazi2019entity}. As we showed in \autoref{tab:tail}, these models perform well over popular entities, but struggle to resolve the tail. \citet{jin2014entity} and \citet{kore50} study disambiguation at the tail, and both rely on phrase-based language models for feature extraction. Unlike our work, they do not fuse type or knowledge graph information for disambiguation.

\paragraph{Disambiguation with Types}
Similar to our work, recent approaches have found that type information can be useful for entity disambiguation~\cite{dredze2010entity,ling2015design,gupta2017entity,raiman2018deeptype,zhu2019latte,chen2020improving}. \citet{dredze2010entity} use predicted coarse-grained types as entity features into a SVM classifier. \citet{chen2020improving} models type information as local context and integrates a BERT contextual embedding into the model from \cite{ganea2017deep}. \citet{raiman2018deeptype} learns its own type systems and performs disambiguation through type prediction alone (essentially capturing the type affordance pattern). \citet{ling2015design} demonstrate that the 112-type FIGER type ontology could improve entity disambiguation, and the LATTE framework~\cite{zhu2019latte} uses multi-task learning to jointly perform type classification and entity disambiguation on biomedical data. \citet{gupta2017entity} adds both an entity-level and mention-level type objective using type embeddings embeddings. We build on these works using fine and coarse-grained entity-level type embeddings and a mention-level type prediction task.

\paragraph{Disambiguation with Knowledge Graphs}
Several recent works have also incorporated (knowledge) graph information through graph embeddings \cite{parravicini2019fast}, co-occurrences in the Wikipedia hyperlink graph \cite{radhakrishnan2018elden}, and the incorporation of latent relation variables \cite{le2018improving} to aid disambiguation. \citet{cetoli2018named} and \citet{mulang2020evaluating} incorporate Wikidata triples as context into entity disambiguation by encoding triples as textual phrases (e.g., ``<subject> <predict> <object>'') to use as additional inputs, along with the original text to disambiguate, into a language model. In \name, the Wikidata connections through the \texttt{KG2Ent} module allow for collective resolution and are not just additional features.

\paragraph{Entity Knowledge in Downstream Tasks}
The works of \citet{knowbert, poerner1911bert, zhang2019ernie, broscheit2020investigating} all try to add entity knowledge into a deep language model to improve downstream natural language task performance. \citet{knowbert, poerner1911bert, zhang2019ernie} incorporate pretrained entity embeddings and finetune either on a the standard masked sequence-to-sequence prediction task or combined with an entity disambiguation/linking task.\footnote{Entity disambiguation refers to when the mentions are pre-detected in text. Entity linking includes the mention detection phase. In \name, we focus on the entity disambiguation task.} On the other hand, \citet{broscheit2020investigating} trains its own entity embeddings. Most works, like \name, see lift from incorporating entity representations in the downstream tasks.

\paragraph{Wikipedia Weak Labelling}
Although uncommon, \citet{broscheit2020investigating, nothman2008transforming, ghaddar2017winer, de2020autoregressive} all apply some heuristic weak labelling techniques to increase link coverage in Wikipedia for either entity disambiguation or named entity recognition. All methods generally rely on finding known surface forms for entities and labelling those in the text. \name is the first to investigate the lift from incorporating weakly labelled Wikipedia data over the tail.

\paragraph{Self-Supervision and the Tail} 

The works of \citet{tata2019itemsuggest}, \citet{chung2019automated}, \citet{ilievski2018systematic}, and \citet{chung2018unknown} all focus on the importance of the tail during inference and the challenges of capturing it during training. They all highlight the data management challenges of monitoring the tail (and other missed slices of data) and improving generalizability. In particular, \citet{ilievski2018systematic} studies the tail in NED and encourages the use of separate head and tail subsets of data. From a broader perspective of natural language systems and generalizability, \citet{ettinger2017towards} highlights that many NLP systems are brittle in the face of tail linguistic patterns. \name builds off this work, investigating the tail with respect to NED and demonstrating the generalizable reasoning patterns over structural resources can aid tail disambiguation.

\section{Conclusion}
\label{sec:conclusion}

We present \name, a state-of-the-art NED system that is explicitly grounded in a principled set of reasoning patterns for disambiguation, defined over entities, types, and knowledge graph relations. The contributions of this work include the characterization and evaluation of core reasoning patterns for disambiguation, a new learning procedure to encourage the model to learn the patterns, and a weak supervision technique to increase utilization of the training data. We find that \name improves over the baseline SotA model by over \yell{40} F1 points on the tail of Wikipedia. Using \name's entity embeddings for a downstream relation extraction task improves performance by \yell{1.0} F1 points, and \name's representations lead to an 8\% lift on highly optimized production tasks at a major technology company. We hope this work inspires future research on improving tail performance by incorporating outside knowledge in deep models.

\vspace{0.3cm}

\begin{spacing}{0.75}
\begin{scriptsize}
\noindent \textbf{Acknowledgements:} We thank Jared Dunnmon, Dan Fu, Karan Goel, Sarah Hooper, Monica Lam, Fred Sala, Nimit Sohoni, and Silei Xu for their valuable feedback and Pallavi Gudipati for help with experiments.  We gratefully acknowledge the support of DARPA under Nos. FA86501827865 (SDH) and FA86501827882 (ASED); NIH under No. U54EB020405 (Mobilize), NSF under Nos. CCF1763315 (Beyond Sparsity), CCF1563078 (Volume to Velocity), and 1937301 (RTML); ONR under No. N000141712266 (Unifying Weak Supervision); the Moore Foundation, NXP, Xilinx, LETI-CEA, Intel, IBM, Microsoft, NEC, Toshiba, TSMC, ARM, Hitachi, BASF, Accenture, Ericsson, Qualcomm, Analog Devices, the Okawa Foundation, American Family Insurance, Google Cloud, Swiss Re, the HAI-AWS Cloud Credits for Research program, and members of the Stanford DAWN project: Teradata, Facebook, Google, Ant Financial, NEC, VMWare, and Infosys. The U.S. Government is authorized to reproduce and distribute reprints for Governmental purposes notwithstanding any copyright notation thereon. Any opinions, findings, and conclusions or recommendations expressed in this material are those of the authors and do not necessarily reflect the views, policies, or endorsements, either expressed or implied, of DARPA, NIH, ONR, or the U.S. Government.
\end{scriptsize}
\end{spacing}

\bibliographystyle{plainnat}
\bibliography{references}

\newpage
\appendix
\section{Extended Model Details}
\label{sec:appendix:model}
We now provide additional details about the model introduced in \autoref{sec:architecture}. We first describe our type prediction module and then describe the added entity positional encoding.
\paragraph{Type Prediction}
To allow the model to further infer the correct types for an entity, especially when the entity does not have a preassigned type, we add a coarse \emph{mention} type prediction task given the mention embedding. Given a mention $m$ and a coarse type embedding matrix $\mathbf{T}$, the task is to assign a coarse type embedding for the mention $m$; i.e., determine $\mathbf{t}_m$. We do so by adding the first and last token of the mention from $\mathbf{W}$ to generate a contextualized mention embedding $\mathbf{m}$. 
We predict the coarse type of the mention $\mathbf{\hat{t}}_m$ by computing
\begin{align*}
    \mathbf{S}_{type} &= \textrm{softmax}(\textrm{MLP}(\mathbf{m})) \\
    \mathbf{\hat{t}}_{m} &= \mathbf{S}_{type}\mathbf{T}
\end{align*}
where $\mathbf{S}_{type}$ generates a distribution over coarse types.
For each entity candidate of $m$, $\mathbf{\hat{t}_m}$ gets concatenated to the other type embedding $\mathbf{t}_e$ before the MLP. This is supervised by minimizing the cross entropy between $\mathbf{S}_{type}$ and the true coarse type for the gold entity, generating a type prediction loss $\mathcal{L}_{type}$. When performing type prediction, our overall loss is $\mathcal{L}_{dis} + \mathcal{L}_{type}$.

\paragraph{Position Encoding}
We need \name to be able to reason over absolute and relative positioning of the words in the sentence and the mentions. For example, in the sentence \emph{``Where is America in Indiana?''}, ``America'' refers to the city in Indiana, not the United States. In the sentence \emph{``Where is Indiana in America?''}, ``America'' refers to the United States. The relative position of ``Indiana'', ``in'', and ``America'' signals the correct answer.

To achieve this signaling, we add the sin positional encoding from \citet{vaswani2017attention} to $\mathbf{E}$ before it is passed to our neural model. Specifically, for mention $m$, we concatenate of the positional encoding of the first and last token of $m$, project the concatenation to dimension $H$, and add it to each of $m$'s $K$ candidates in $\mathbf{E}$. As we use BERT word embeddings for $W$, the positional encoding is already added to words in the sentence.

\section{Extended Results}
\label{sec:appendix:results}
We now give the details of our experimental setup and training. We then give extended results over the regularization scheme and model ablations. Lastly, we extend our error analysis to validate \name's ability to reason over the four patterns.
\subsection{Evaluation Data}
\paragraph{Wikipedia Datasets}
We use two main datasets to evaluate \name.
\begin{itemize}[leftmargin=*, itemsep=2pt, topsep=2pt]
    \item{\textbf{Wikipedia}}: we use the November 2019 dump of Wikipedia to train \name. We use the set of entities that are linked to in Wikipedia for a total of \yell{3.3}M entities. After weak labelling, we have a total of \yell{5.7}M sentences.
    \item{\textbf{Wikipedia Subset}}: we use a subset of Wikipedia for our micro ablation experiments over regularization parameters. We generate this subset by taking all sentences where at least one mention is a mention from the KORE50 disambiguation benchmark. Our set of entities is all entities and entity candidates referred to by mentions in this subset of sentences. We have a total of 370,000 entities and 520,000 sentences.
\end{itemize}
For our Wikipedia experiments, we use a 80/10/10 train/test/dev split by Wikipedia pages, meaning all sentences for a single Wikipedia page get placed into one of the splits. For our benchmark model, we use a 96/2/2 train/test/dev split over sentences to allow our model to learn as much as possible from Wikipedia for our benchmark tasks.

\paragraph{Benchmark Datasets}
We use three benchmark NED datasets. Following standard procedure~\cite{ganea2017deep}, we only consider mentions whose linked entities appear in Wikipedia. The datasets are summarized as follows:
\begin{itemize}[leftmargin=*, itemsep=2pt, topsep=2pt]
    \item{\textbf{KORE50}}: KORE50~\cite{kore50} represents difficult-to-disambiguate sentences and contains $144$ mentions to disambiguate. Note, as of the Nov 2019 Wikipedia dump, one mention in the 144 does not have a Wikipedia page. Although it is standard to remove mentions that do not link to an entity in Wikipedia, to be comparable to other methods, we measure with 144 mentions, not 143.
    \item{\textbf{RSS500}}: RSS500~\cite{rss500} is a dataset of news sentences and contains $520$ mentions (4 of the mentions did not have entities in $\mathcal{E}$).
    \item{\textbf{AIDA CoNLL-YAGO}}: AIDA CoNLL-YAGO~\cite{hoffart2011robust} is a document-based news dataset containing $4,485$ test mentions, $4,791$ validation set mentions, and $18,541$ training mentions. As \name is a sentence-level NED system, we create sentences from documents following the technique from \citet{fevry2020empirical} where we concatenate the title of the document to the beginning of each sentence.
\end{itemize}
To improve the quality of annotated mention boundaries in the benchmarks, we follow the technique of~\citet{Phan2019PairLinkingFC} and allow for mention boundary expansion using a standard off-the-shelf NER tagger.\footnote{We use the spaCy NER tagger from \url{https://spacy.io/}} For candidate generation, as aliases may not exist in $\Gamma$, we gather possible candidates by looking at n-grams in descending order of length and determine the top 30 by measuring the similarity of the proper nouns in the example sentence to each candidate's Wikipedia page text.

\paragraph{Structural Resources}
The last source of input data to \name is the structural resources of types and knowledge graph relations. We extract relations from Wikidata knowledge graph triples. For our pairwise KG adjacency matrix used in \texttt{KG2Ent}, we require the subject and object to be in $\mathcal{E}$. For our relation embeddings, we only require the subject be in $\mathcal{E}$ as our goal is to extract all relations an entity participates in independent of the other entities in the sentence. We have a total of 1,197 relations.

We use two different type sources to assign types to entities---Wikidata types and HYENA types~\cite{hyena}---and use coarse HYENA types for type prediction. The Wikidata types are generated from Wikidata's ``instance of'', ``subclass of'', and ``occupation'' relationships.  The ``occupation" types are used to improve disambiguation of people, which otherwise only receive ``human" types in Wikidata. We filter the set of Wikidata types to be only those occurring 100 or more times in Wikipedia, leaving \yell{27}K Wikidata types in total. The HYENA type hierarchy has 505 types derived from the YAGO type hierarchy. 
We also use the coarsest HYENA type for an entity as the gold type for type prediction. There are $5$ coarse HYENA types of person, location, organization, artifact, event, and miscellaneous.

\subsection{Training Details}
\paragraph{Model Parameters}
We run three separate models of \name: two on our full Wikipedia data (one for the ablation and one for the benchmarks) and one on our micro data. For all models we use 30 candidates for each mention and incorporate the structural resources discussed above. We set $T = 3$ and $R = 50$ for the number of types and relations assigned to each entity.

For the models trained on our full Wikipedia data, we set the hidden dimension to $512$, the dimension of $\mathbf{u}$ to $256$, and the dimension of all other type and relation embeddings to $128$. For our models trained on our micro dataset, we set the hidden dimension to $256$, the dimension of $\mathbf{u}$ to $256$, and the dimension of all other type and relation embeddings to $128$.

The final differences to discuss are between the benchmark model and ablation model over all of Wikipedia. To make the best performing model for benchmarks, we add two additional additional features we found improved performance:

\begin{itemize}[leftmargin=*, itemsep=2pt, topsep=2pt]
\item We use an additional \texttt{KG2Ent} module in addition to an adjacency matrix indicating if two entities are connected in Wikidata. We add a matrix containing the log of the number of times two entities occur in a sentence together in Wikipedia. If they co-occur less than 10 times, the weight is 0. We found this helped the model better learn entity co-occurrences from Wikipedia.

\item We allow our model to use additional entity-based features to be concatenated into our final $\mathbf{E}$ matrix. We add two features. The first is the average BERT WordPiece embeddings of the title of an entity. This is similar to improving tail generalization by embedding a word definition in word sense disambiguation~\cite{blevins2020moving}. 
This allows the model to better capture textual cues indicating the correct entity. We also append a 1-dimensional feature of how many other entities in the sentence appear on the entity's Wikipedia page. This increases the likelihood of an entity that has more connection to other candidates in the sentence. We empirically find that using the page co-occurrences as an entity feature rather than as a \texttt{KG2Ent} module performs similarly and reduces the runtime.
\end{itemize}

Further, our benchmark model uses a fixed regularization scheme of $80\%$ which did not hurt benchmark performance and training was marginally faster than the inverse popularity scheme. We did not use these features for ablations as we wanted a clean study of the model components as described in \autoref{sec:architecture} with respect to the reasoning patterns.

\paragraph{Training}
We initialize all entity embeddings to the same vector to reduce the impact of noise from unseen entities receiving different random embeddings. We use the Adam optimizer~\cite{adam} with a learning rate of $0.0001$ and a dropout of 0.1 in all feedforward layers, 16 heads in our attention modules, and we freeze the BERT encoder stack. Note for the NED-Base model, we do not freeze the encoder stack to be consistent with~\citet{fevry2020empirical}.

For the models trained on all of Wikipedia, we use a batch size of 512 and train for 1 epoch for the benchmark model and 2 epochs for the ablation models on 8 NVIDIA V100 GPUs. For our micro data model, we use a batch size of 96 and train for 8 epochs on a NVIDIA P100 GPU.
 
\subsection{Extended Ablation Results}
\begin{table}[]
\begin{center}
 \caption{(top) We compare \name to a BERT-based NED baseline (NED-Base) on validation sets of our micro Wikipedia dataset and ablate \name by only training with entity, type, or knowledge graph data. We further ablate (bottom 8 rows) the regularization schemes for the entity embeddings for \name.}
    \vspace{3mm}
    \label{tab:micro_ablation}
\begin{tabular}{lcccc}
\toprule
 Model           & All Entities & Torso Entities & Tail Entities & Unseen Entities \\
            \midrule

NED-Base    & 90.2         & 91.6           & 50.5          & 21.5            \\

\midrule
\name (Ent-only) & 89.1         & 89.0           & 48.3          & 15.5            \\
\name (Type-only)   & 91.6         & 90.4           & 65.9          & 56.8            \\
\name (KG-only)    &    91.8          &  90.8              &   65.3     &  58.6        \\
\name ($p(e) = 0\%$)    &    92.5          &  92.3              &   67.7     &  48.6        \\
\name ($p(e) = 20\%$)    &    92.8          &  92.5              &   68.9     &  52.5        \\
\name ($p(e) = 50\%$)    &    \textbf{92.9} &  \textbf{92.7}    &   70.1     &  57.7        \\
\name ($p(e) = 80\%$)    & 92.8         & 92.2                & 69.5          & 59.9            \\
\name (InvPopLog)    &    92.7          &  91.9              &   69.7     &  61.1        \\
\name (InvPopPow)    &    92.8          &  92.3              &   \textbf{70.5}     &  \textbf{62.2}        \\
\name (InvPopLin)    &    92.6          &  91.8              &   69.7     &  61.0        \\
\name (PopPow)       &    \textbf{92.9}          &  92.5              &   68.9     &  52.4        \\
\midrule
\# Mentions & 96,237      & 37,077         & 11,087         & 2,810  
\\
\bottomrule
\end{tabular}
\end{center}
\end{table}
\paragraph{Ablation Model Size} \autoref{tab:model_sizes} reports the model sizes of each of the five ablation models from \autoref{tab:tail}. As we finetuned the BERT language model in NED-Base (to be consistent with \citet{fevry2020empirical}) but do not do so in \name, we do not count the BERT parameters in our reported sizes to be comparable.

\begin{table}[h]
\begin{center}
\caption{We report the model sizes in MB of each of the five ablation models: NED-Base, \name, \name (Ent-Only), \name (KG-Only), and \name (Type-Only).}
\vspace{3mm}
\label{tab:model_sizes}
\begin{tabular}{llllll}
\toprule
 Model           & NED-Base & \name & Ent-Only & Type-Only & KG-Only \\ \midrule
 Embedding Size (MB) & 5,186 & 5,201 & 5,186 & 13 & 1 \\ 
 Network Size (MB) & 4 & 39 & 35 & 38 & 34 \\
 Total Size (MB) & 5,190 & 5,240 & 5,221 & 51 & 35 \\
\bottomrule
\end{tabular}
\end{center}
\end{table}


\paragraph{Regularization} We now present the extended results of our regularization and weak labelling ablations over our representative micro dataset. \autoref{tab:micro_ablation} gives full ablations over a variety of regularization techniques. As in \autoref{tab:tail}, we include results from models using only the entity, type, or relation information, in addition to the BERT and \name models. 

We report the results of inverse popularity regularization based on three different functions that map the the curve of entity counts in training to the regularization value. For each function, we fix that entities with a frequency 1 receive a regularization value of $0.95$ while entities with a frequency of 10,000 receive a value of $0.05$ and assign intermediate values to generate a linear, logarithmic, and power curve that applies less regularization for more popular entities. The regularization reported in \autoref{tab:regularization} uses a power law function ($f(x) = 0.95(x^{-0.32})$). We also report in \autoref{tab:micro_ablation} a linear function ($f(x) = -0.00009x + 0.9501$) and a logarithmic function ($f(x) = -0.097\log(x) + 0.96$). Each regularization function is set to a range of 0.05 to 0.95. We leave it as future work to explore other varied regularization functions.

\autoref{tab:micro_ablation} shows similar trends as reported in \autoref{sec:evaluation} that \name with all sources of information and the power law inverse regularization performs best over the tail and the unseen entities. We do see that the model trained with a fixed regularization of 0.5 performs marginally better on the torso and over all entities, likely because this involves less regularization over those entity embeddings, allowing it to better leverage the memorized entity patterns, while also leveraging some type and relational information (as shown by its improved performance over a lower fixed regularization). This model, however, performs 4.5 F1 points worse over unseen entities than the best model.

\paragraph{Weak Labeling} \autoref{tab:micro_ablation_wl} shows \name's results with the inverse power law regularization with and without weak labelling. For this ablation, we define our set of torso, tail, and unseen entities by counting entity occurrence {\em before} weak labelling to have a better understanding as to the lift from adding weak labelling (rather than the drop without it).

\begin{table}[]
\begin{center}
\caption{We report \name trained with versus without weak labelling on our micro Wikipedia dataset. The slices defined by gold anchor counts (pre-weak labelling). We use the InvPopPow regularization for both.}
\vspace{3mm}
\label{tab:micro_ablation_wl}
\begin{tabular}{lcccc}
\toprule
 Model           & All Entities & Torso Entities & Tail Entities & Unseen Entities \\
\midrule
\name             &    92.8          &  92.6              &   \textbf{70.5}     &  \textbf{63.3}        \\
\name (No WL)    &    \textbf{93.3}   &  \textbf{93.1}      &   70.2     &  60.7        \\
\midrule
\# Mentions & 96,237      & 36,904         & 11,541         & 3,146  
\\
\bottomrule
\end{tabular}
\end{center}
\end{table}
We see that weak labelling provides a lift of 2.6 F1 points over unseen entities and 0.3 F1 points over tail entities. Surprisingly, without weak labeling, \name performs 0.5 F1 points better on torso entities. We hypothesize this occurs because the noisy weakly labels increase the amount available signals for \name to learn consistency patterns for rarer entities---noisy signals are better than no signals---however, popular entities have enough less-noisy gold labels in the training data, so the noise from weak labels may create conflicting signals that hurt performance. 

To validate this hypothesis, we see that overall, counting both true and weakly labelled mentions, 4\% of mentions without weak labeling share the same types as at least one other mention in the sentence while 14\% of mentions with weak labelling do. Our model predicts a consistent answer only 4\% of the time without weak labeling compared to 13\% of the time with weak labeling. Note this is a slightly higher coverage numbers than reported in \autoref{sec:analysis} as we are using a weaker form of consistency---two mentions in the sentence share the same type independent of position and ordering---and are including weakly labelled mentions. This indicates consistency is a significantly more useful pattern with weak labelling, and our model predicts more consistent answers with weak labelling than without. Over the torso with weak labelling, we find that 14\% of the errors across all mentions (weak labelled and anchor) are when \name uses consistency reasoning, but the correct answer does not follow the consistency pattern. Without weak labelling, only 5\% of the errors are due to consistency.

\section{Extended Downstream Details}
\label{sec:appendix:downstream}
We now provide additional details of our SotA TACRED model, which uses \name embeddings.
\paragraph{Input} We use the revisited TACRED dataset \cite{Alt2020TACREDRA}: each example includes text and subject and object positions in the text. The task involves extracting the relation between the subject and object. There are 41 potential relations as well as a ``no relation'' option. The other features we use as inputs are NER, POS tags, and contextual \name embeddings for entities that \name disambiguates in the sentence.

\paragraph{\name Model} As TACRED does not come with existing mention boundaries, we perform mention extraction by searching over n-grams, from longest to shortest, in the sentence and extract those that are known mentions in \name's candidate maps. We use the same \name model from our ablations with entity, KG, and type information except with the addition of fine-tuned BERT word embeddings. For efficiency, we train on a subset of Wikipedia training data relevant to TACRED. To obtain the relevant subset, we take Wikipedia sentences containing entities extracted during candidate generation from a uniform sample of TACRED data; i.e., entities in the candidate lists of detected mentions from a uniform TACRED sample. The contextualized entity embeddings from \name over all TACRED are fed to the downstream model. 

\paragraph{Downstream Model} We first use standard SpanBERT-Large embeddings to encode the input text, concatenate the contextual \name embeddings with the SpanBERT embeddings, and then pass this through four transformer layers. We then calculate the cross-entropy loss and apply a softmax for scoring. We freeze the \name embeddings and fine-tune the SpanBERT embeddings. We use the following hyperparameters: the learning rate is 0.00002, batch size is 8, gradient accummulation is 6, number of epochs is 10, L2 regularization is 0.008, warm-up percentage is 0.1, and maximum sequence length is 128. We train with a NVIDIA Tesla V100 GPU.

\paragraph{Extended Results}
We study the model performance as a function of the signals provided by \name. In \autoref{tab:tacred_full}, we show that on slices with above-median numbers of Bootleg entity, relation, and type signal counts identified in the TACRED example, the relative gap between BERT and Bootleg errors is larger on the slice above, than below, the median by 1.10x, 4.67x, and 1.35x respectively. In \autoref{tab:tacred_subjobj} we show the relative error rates from the \name and baseline SpanBERT models for the slices where \name provides an entity, relation, or type signal for the TACRED example's subject or object. On the slice of these signals are respectively present, the baseline model performs 1.20x, 1.18x, and 1.20x worse than the \name TACRED model. These results indicate that the knowledge representations from \name successfully transfer useful information to the downstream model.

\begin{table}[h!]
\begin{center}
\captionof{table}{We rank TACRED examples by the proportion of words that receive \name embedding features where: \name disambiguates an entity, leverages Wikidata relations for the embedding, and leverages Wikidata types for the embedding. We take examples where the proportion is greater than 0. For each of these three slices, we report the gap between the SpanBERT model and Bootleg model's error rates on the examples with above-median proportion (more \name signal) relative to the below-median proportion (less \name signal). With more \name information, we see the improvement our SotA model provides over SpanBERT increases.}
\begin{tabular}{lccc}
\toprule
\name Signal & \# Examples with the Signal & Gap Above/Below Median\\
            \midrule
 Entity   &    15323 & 1.10           \\
 Relation   &  5400 & 4.67          \\
 Type   & 15509 & 1.35          \\
\bottomrule
\end{tabular}
\label{tab:tacred_full}
\end{center}
\end{table}

\begin{table}[h!]
\begin{center}
\captionof{table}{We compute the error rate of SpanBERT relative to our \name downstream model for three slices of TACRED data where respectively \name disambiguates the subject and/or object, \name leverages Wikidata relations for the embedding of the subject and object pair, and \name leverages Wikidata types for the embedding of the subject and/or object in the example.}
\begin{tabular}{lcc}
\toprule
 Subject-Object Signal & \# Examples & BERT/\name Error Rate \\
            \midrule
Entity   & 12621  & 1.20           \\
Relation    &  542 & 1.18           \\
Obj Type     &  12044 & 1.20        \\
\bottomrule
\end{tabular}
\label{tab:tacred_subjobj}
\end{center}
\end{table}


\section{Extended Error Analysis}
\label{sec:appendix:error}
\subsection{Reasoning Patterns}
Here we provide additional details about the distributions of types and relations in the data.

\paragraph{Distinct Tails} Like entities, types and relations also have tail distributions. For example, types such as ``country'' and ``film'' appear 2.7M and 800k times respectively, while types such as ``quasiregular polyhedron'' and ``hospital-acquired infection'' appear once each in our Wikipedia training data. Meanwhile, relations such as ``occupation'' and ``educated at'' appear 35M and 16M times respectively, while relations such as ``positive diagnostic predictor'' and ``author of afterword'' appear 7 times respectively in the Wikipedia training data. However we find that the entity-, relation-, and type-tails are distinct: 88\% of the tail-entities by entity-count have Wikidata types that are non-tail types and 90\% of the tail-entities by entity-count have non-tail relations.\footnote{Similar to tail-entities, tail-types and tail-relations are defined as those appearing 1-10 times in the training data.} For example, the head Wikidata type \yell{``country''} contains rare entities \yell{``Palaú'' and ``Belgium–France border''}. 

We observe that \name significantly improves tail performance over each of the tails. We rank the Wikidata types and relations by the number of occurrences in the training data and study the lift from \name as a function of the number of times the signal appears during training. \name performs an \yell{9.4 F1} and \yell{20.3 F1} points better than the NED-Base baseline for examples with gold types appearing more and less than the median number of times during training respectively. \name provides a a \yell{7.8 F1 points} and \yell{13.7 F1} points better than the baseline for examples with gold relations appearing more and less than the median number of times during training respectively. These results indicate that \name excels on the tails of types and relations as well.  

\begin{figure*}[t]
    \centering
    \includegraphics[width=0.49\linewidth]{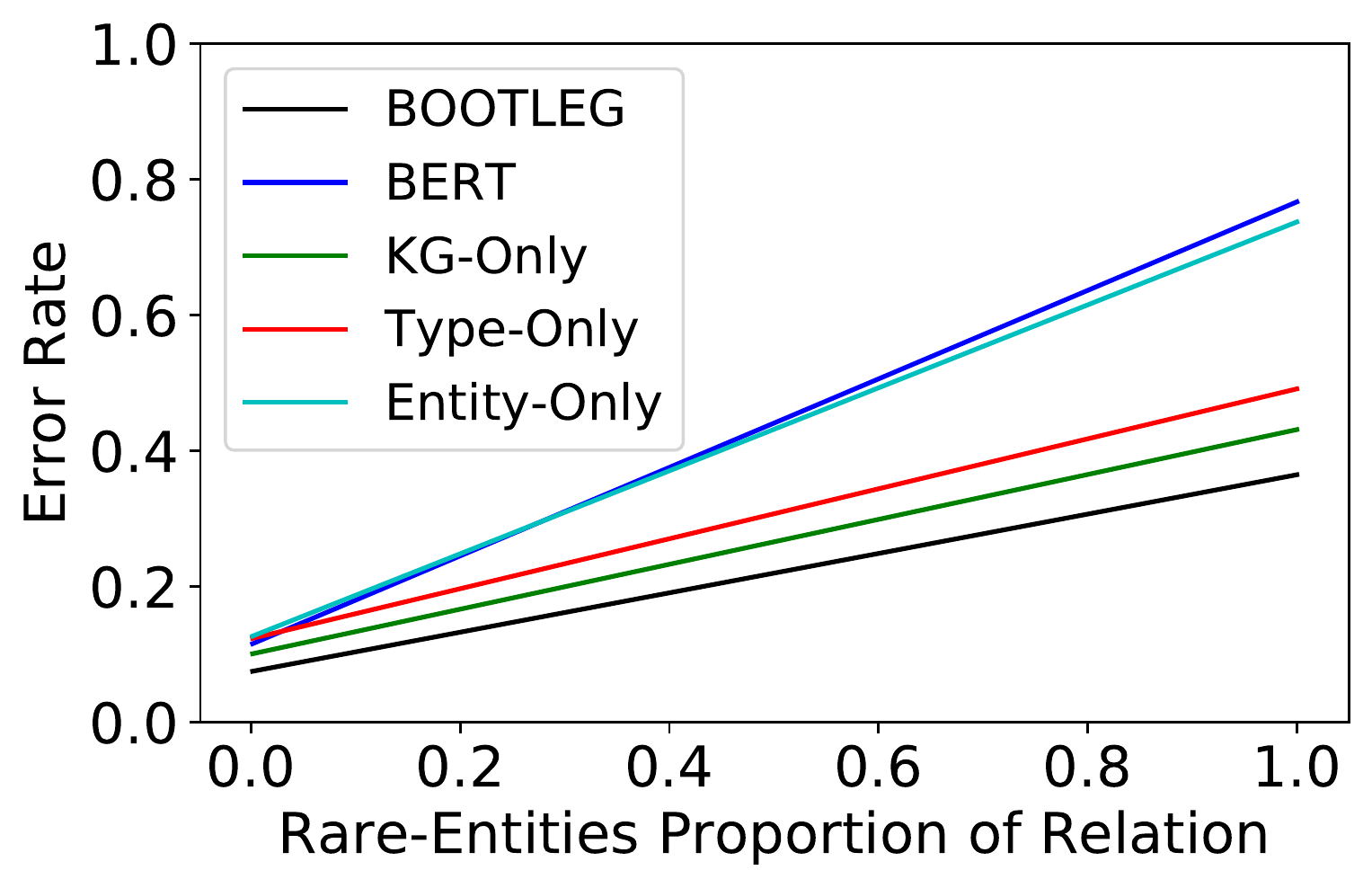}
    \includegraphics[width=0.49\linewidth]{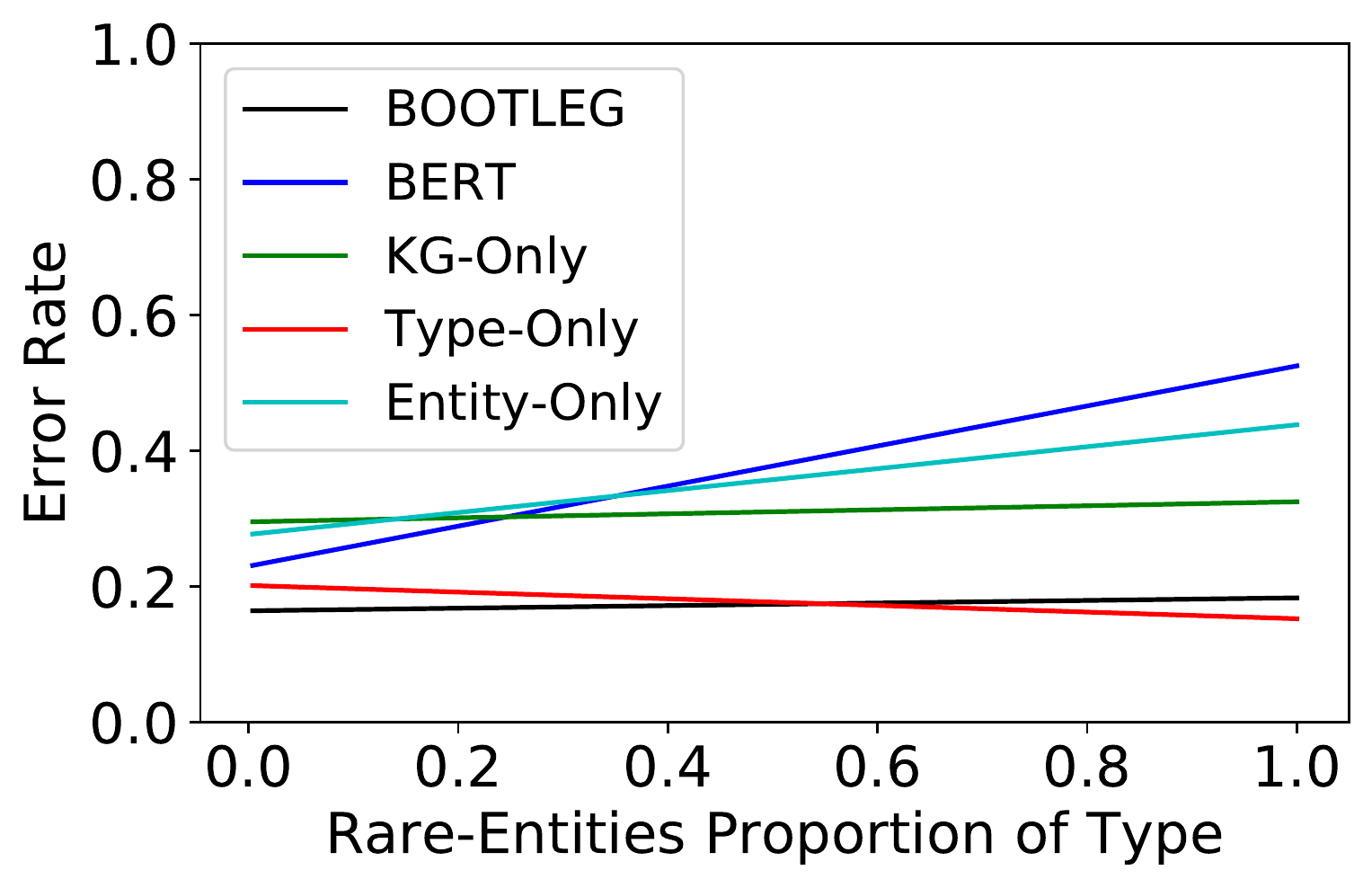}
    \caption{For all the entities of a particular type or relation, we calculate the percentage of rare entities (tails and toes entities). We show the error rate on the Wikipedia validation set as a function of the rare-proportion of entities of a given (Left) relation or (Right) type appearing in the validation set.}
    \label{fig:rare_prop}
\end{figure*}

Next, ranking the Wikidata types and relations by the proportion of comprised rare (tail and unseen) entities, we further find that \name provides the lowest error rates across types and relations, regardless of the proportion of rare entities, while the baseline and Entity-Only models give relatively larger error rates as the proportion of rare entities increases (\autoref{fig:rare_prop}). The trend for types is flat as the proportion of rare entities increases, while the trend for relations is upwards sloping. 
These results indicate that \name is better able to transfer the patterns learned from one entity to other entities that share its same types and relations. The improvement from \name over the baseline increases as the rare-proportion increases, indicating that \name is able to efficiently transfer knowledge even when the type or relation category contains none or few popular entities. 

\paragraph{Type Affordance} 
 For the type affordance pattern, we find that the TF-IDF keywords provide high coverage over the examples containing the gold type: 88\% of examples where the gold entity has a particular type contain an affordance keyword for that type. An example of a type with full coverage by the affordance keywords is ``café'', with keywords such as ``coffee'', ``Starbucks'', and ``Internet''; in each of the 77 times an entity of the type ``cafe'' appears in the validation set, an affordance keyword is present. Types with low coverage in the validation set by affordance keywords tend to be the rare types: for the types with coverage less than 50\%, such as ``dietician'' or ``chess official'', the median number of occurrences in the validation set is 1. This supports the need for knowledge signals with distinct tails, which can be assembled to together address the rare examples.

\end{document}